# A Quantum-inspired Hybrid Swarm Intelligence and Decision-Making for Multi-Criteria ADAS Calibration

Sanjai Pathak
*Amity University Uttar Pradesh*
Noida, India
pathak.sanjai@gmail.com

Ashish Mani
*Amity University Uttar Pradesh*
Noida, India
amani@amity.edu

Amlan Chatterjee
*California State University*
Dominguez Hills Carson, CA, USA
achatterjee@csudh.edu

*Abstract* - The tuning of Advanced Driver Assistance Systems (ADAS) involves resolving trade-offs among several competing objectives, including operational safety, system responsiveness, energy usage, and passenger comfort. This work introduces a novel optimization framework based on Quantum-Inspired Hybrid Swarm Intelligence (QiHSI), in which quantum-inspired mechanisms are embedded within a multi-objective salp swarm optimization process to strengthen global search capability and preserve population diversity in complex, high-dimensional decision spaces. In addition, a decision-maker-in-the-loop strategy is incorporated to incorporate adaptive expert guidance, enabling the optimization process to respond dynamically to changing design priorities and system constraints. The effectiveness of QiHSI is assessed using established multi-objective benchmark problems as well as a practical ADAS calibration scenario. Experimental comparisons with several state-of-the-art evolutionary and swarm-based algorithms, including MSSA, MOPSO, MOEA/D, SPEA2, NSGA-III, and RVEA, show that the proposed method consistently produces well-distributed Pareto-optimal solutions with faster convergence and improved adaptability. These findings demonstrate that QiHSI offers a reliable and scalable approach for intelligent ADAS calibration, supporting the development of more responsive, efficient, and safety-oriented autonomous driving technologies.

## I. Introduction

Advanced Driver Assistance Systems (ADAS) are fundamental to modern automotive engineering, enhancing safety, performance, energy efficiency, and comfort through intelligent control and sensing technologies. The calibration of ADAS involves tuning multiple interdependent parameters where improving one criterion often compromises another, for instance, maximizing fuel efficiency may reduce braking responsiveness, while improving comfort may degrade safety thresholds. Consequently, ADAS calibration represents a challenging multi-objective optimization problem [1], [2]. Traditional optimization methods have been used for this task; however, the high dimensionality, nonlinear parameter interdependencies, and conflicting objectives make achieving balanced trade-offs computationally demanding [3], [4].

Metaheuristic algorithms, which rely on stochastic exploration rather than gradient information, have been widely applied to solve complex, black-box, and multimodal optimization problems across engineering, artificial intelligence, and control domains [5]–[7]. Techniques such as Multi-Objective Evolutionary Algorithms (MOEA) [8], Genetic Algorithms (GA) [9], Particle Swarm Optimization (PSO) [10], Ant Colony Optimization (ACO) [11], and Differential Evolution (DE) [12] provide effective global search capabilities but often suffer from premature convergence, loss of diversity, and inefficient handling of variable dependencies, especially in high-dimensional ADAS calibration spaces. Recent developments in swarm and evolutionary optimization have sought to overcome these limitations. Well-established algorithms such as NSGA-II [14], MOPSO [15], SPEA2 [16], and MOEA/D [17] have achieved strong results in multi-objective optimization; however, they typically rely on static population structures and fixed variation operators, which limit adaptability under dynamically changing constraints [6], [18]. Advanced frameworks such as NSGA-III [39] and RVEA [40] have improved reference-vector-based diversity preservation but still face difficulties in real-time adaptive optimization where objective weights or constraints evolve according to system feedback.

Within automotive applications, optimization algorithms like PSO [18], [19], ACO [20], and the Multi-Objective Salp Swarm Algorithm (MSSA) [5], [21] have been used for calibration tasks including sensor fusion, controller tuning, and energy efficiency improvement. While these methods achieve satisfactory convergence, they often stagnate in local optima and exhibit poor scalability in large-scale, multi-criteria environments [22]–[25]. Hybrid strategies that combine swarm and evolutionary mechanisms such as PSO with genetic operators, ACO with pheromone-based reinforcement, or cooperative model of swarm intelligence with differential evolution offer partial improvements but remain limited in handling dynamic human feedback or evolving system priorities [18], [20], [41].

Recent studies have shown that optimization strategies inspired by quantum computing concepts offer a powerful mechanism for enhancing search behavior in complex optimization problems. By exploiting probabilistic state representations and correlated variable interactions—analogous to quantum superposition, entanglement, and

tunneling—these methods enable broader exploration of the solution space and reduce the likelihood of stagnation at local optima [26], [27]. Several approaches, such as the Quantum-Inspired Genetic Algorithm (QGA) [28], Quantum Particle Swarm Optimization (QPSO) [29], and Quantum-Inspired Salp Swarm Algorithm (QSSA) [30], [31], have reported notable gains in convergence efficiency and population diversity, particularly in dynamic and high-dimensional optimization scenarios. However, their integration into practical automotive calibration remains limited.

Motivated by these challenges, this paper introduces the Quantum-Inspired Hybrid Swarm Intelligence (QiHSI) framework for multi-criteria ADAS calibration. QiHSI integrates the Multi-Objective Salp Swarm Algorithm (MSSA) with quantum superposition and entanglement operators to enhance exploration and maintain solution diversity while mitigating stagnation in complex search landscapes. Additionally, a Decision-Maker-in-the-Loop (DMiL) mechanism enables real-time expert feedback to dynamically adjust optimization priorities in response to evolving system constraints. Through this integration, QiHSI delivers a scalable, adaptive, and human-aligned optimization framework for next-generation autonomous and semi-autonomous driving systems. The primary contributions of this work are summarized as follows:

1. A novel Quantum-Inspired Hybrid Swarm Intelligence algorithm integrating quantum operators with MSSA to enhance exploration, diversity, and convergence efficiency.
2. Introduction of a dynamic human-feedback mechanism for real-time adaptation of objective priorities during optimization, referred as Decision-Maker-in-the-Loop (DMiL).
3. Extensive benchmarking against MSSA, MOPSO, MOEA/D, SPEA2, NSGA-III, and RVEA across standard multi-objective test suites (ZDT, UF) and a real-world multi-criteria ADAS calibration problem, demonstrating superior Pareto-front convergence, robustness, and adaptability.

The rest of this manuscript is organized as follows. The formulation of the multi-criteria ADAS calibration task as a multi-objective optimization problem is introduced in Section II. Section III explains the proposed Quantum-Inspired Hybrid Swarm Intelligence (QiHSI) framework along with its core algorithmic mechanisms. Experimental evaluation and detailed performance analysis on standard benchmark problems and a practical ADAS calibration case are presented in Section IV. The paper is concluded in Section V with a summary of the main contributions and potential avenues for future work.

## II. Problem Formulation

The calibration of Advanced Driver Assistance Systems (ADAS) is a high-dimensional multi-objective optimization task that aims to balance conflicting design criteria such as safety, performance, energy efficiency, and comfort [2], [32]. Each objective reflects a distinct system requirement: safety corresponds to minimizing collision probability, performance relates to response time, energy efficiency addresses actuator and sensor power consumption, and comfort represents ride smoothness [14], [33], [35]. The interdependence among calibration parameters such as radar sensitivity, controller gain, and braking response creates nonlinear trade-offs, making manual or trial-and-error calibration approaches infeasible [5], [33].

Formally, the ADAS calibration can be represented as a constrained multi-objective optimization problem:

$$\textbf{Minimize } \textbf{\textit{F}}(x) = [f_1(x), f_2(x), f_3(x), \ldots, \ldots f_m(x)]^T$$

subject to:

$$x \in X, \quad X = \{x \in \mathbb{R}^n \mid g_j(x) \leq 0, j = 1, \ldots, k\}$$

where $x = (x_1, x_2, \ldots, x_n)$ denotes the vector of calibration parameters (e.g., sensor angles, controller thresholds, and gain settings), and each $f_i(x)$ corresponds to a performance criterion such as: $f_1(x)$: Safety (collision avoidance probability) [35]; $f_2(x)$: Performance (system response latency) [33]; $f_3(x)$: Energy efficiency (power consumption) [6]; $f_4(x)$: Comfort (ride smoothness) [14].

The feasible solution space is bounded by both physical and regulatory constraints, reflecting limits on hardware parameters and safety compliance requirements. Because these objectives are inherently conflicting, no single global optimum exists. Instead, the solution is represented by a Pareto-optimal set, where no objective can be improved without degrading another. Automotive engineers or control designers act as decision-makers, selecting a final operating point from this Pareto front based on safety priorities, regulatory standards, or vehicle-specific design trade-offs. Furthermore, practical ADAS calibration introduces additional challenges such as environmental uncertainty, sensor noise, and real-time computational limitations [14], [33]. Parameter coupling (e.g., between braking force and steering correction) often leads to interdependencies that hinder independent optimization of variables. Existing algorithms such as MOPSO, MOEA/D, and MSSA have been applied to address these issues, but they frequently experience loss of diversity and premature convergence in such complex, high-dimensional search spaces [33], [36].

To overcome these challenges, this study proposes the Quantum-Inspired Hybrid Swarm Intelligence (QiHSI) framework, which integrates quantum operators including superposition and entanglement with the Multi-Objective Salp Swarm Algorithm (MSSA) to enhance exploration and maintain diversity [26]. Additionally, the framework embeds a Decision-Maker-in-the-Loop (DMiL) component that dynamically adjusts objective weights or dominance relations based on expert feedback, ensuring adaptability under evolving operational conditions. This hybridization enables QiHSI to efficiently approximate high-quality Pareto fronts and align optimized solutions with real-world expert-driven calibration decisions.

## III. Quantum-inspired Hybrid Swarm Intelligence Framework

The proposed Quantum-Inspired Hybrid Swarm Intelligence (QiHSI) framework formulates ADAS calibration as a dynamic multi-objective optimization process, where safety, comfort, and energy efficiency are simultaneously optimized under evolving driving and environmental conditions. The system operates in discrete global optimization cycles $t_1, t_2, \ldots, t_T$, each corresponding to an adaptive calibration phase involving perception, computation, and decision-making.

### A. System Workflow

At each optimization cycle, QiHSI performs five coordinated processes:

1. *Perception and Data Acquisition:* The vehicle's sensors (camera, radar, LiDAR, ultrasonic) collect environmental and operational data, such as lane markings, road curvature, vehicle speed, and obstacle proximity.
2. *Feature Extraction and Fusion:* Multisensory fusion algorithms integrate these inputs to form a structured situational model, enhancing reliability through noise reduction and redundancy management.
3. *Optimization and Calibration Adjustment:* The QiHSI algorithm updates the calibration parameters using hybrid swarm intelligence enhanced with quantum-inspired search mechanisms. This process seeks to optimize multi-criteria objectives including the Safety Index (SI), Energy Efficiency Metric (EEM), and Comfort Performance Score (CPS).
4. *Decision-Maker-in-the-Loop (DMiL) Feedback:* An expert or adaptive decision model dynamically adjusts the weighting of objectives in response to changing road, driver, or environmental conditions, allowing human reasoning to guide the optimization trajectory.
5. *Evaluation and Convergence Check:* The algorithm evaluates whether safety and performance thresholds are met. If criteria are satisfied, the calibration parameters are stored; otherwise, the system continues iterating toward convergence.

This iterative feedback-driven process enables QiHSI to adapt ADAS calibration dynamically to changing environmental, hardware, and driver contexts, ensuring sustained system performance across conditions.

### B. Formulation of Quantum-Inspired Operators -

QiHSI extends the classical Multi-Objective Salp Swarm Algorithm (MSSA) by embedding quantum-inspired operators to enhance exploration and maintain inter-parameter correlation. The quantum superposition operator enables the leader salp to probabilistically combine its position with the best global solution, expanding the search space:

$$q_j = \alpha\, x_{\text{best},j} + (1-\alpha)\, x_{i,j}, \alpha \in [0,1] \quad (1)$$

where $x_{\text{best},j}$ denotes the best non-dominated solution (food source) from repository $R$. To further refine exploration, a quantum entanglement operator introduces correlated perturbations using a rotation transformation:

$$q'_j = q_j \cos(\phi) - \beta \sin(\theta) \quad (2)$$

This interaction models covariance among objectives, directing search steps along correlated improvement directions and preventing premature convergence. The leader's position update then follows:

$$x_{i,j}^{\text{new}} = F_j + c_1[(X_{\max,j} - X_{\min,j})q'_j + X_{\min,j}], \quad (3)$$

where $F_j$ represents the selected food source from $R$, and $c_1 = 2e^{-(4t/T)^2}$ is a convergence control parameter that gradually reduces exploration over iterations.

Follower salps update their positions according to the classical chain model:

$$x_i^{\text{new}} = \frac{1}{2}(x_i^{\text{old}} + x_{i-1}^{\text{new}}), \quad (4)$$

ensuring smooth propagation of information across the swarm while maintaining stability in high-dimensional spaces.

### C. Repository Maintenance and Pareto Diversity Preservation

The repository $R$ stores the non-dominated solutions discovered during evolution and is updated iteratively to preserve diversity and convergence stability. The update rules are as follows:

- Newly generated non-dominated solutions are added to $R$.
- Dominated members are removed and replaced by superior solutions.
- When $|R|$ exceeds its capacity, crowding-distance-based pruning removes densely clustered members to maintain uniform Pareto-front coverage.

Leader salps select their food sources from $R$ using crowding-distance-based roulette selection, favoring less crowded solutions to promote a well-distributed Pareto front and prevent local stagnation.

### D. Decision-Maker-in-the-Loop (DMiL) Integration

QiHSI incorporates a Decision-Maker-in-the-Loop (DMiL) module to introduce adaptive human feedback during optimization. Every $\tau$ iterations, expert feedback or a simulated decision model modifies the objective weights $w_i$ based on safety, comfort, or performance observations. The adaptive weight update follows:

$$w(t+1) = (1-\gamma)\, w(t) + \gamma\, w_{\text{expert}}(t), \quad (5)$$

where $\gamma \in [0,1]$ controls the feedback influence. This enables the search to adaptively prioritize objectives according to real-world conditions or expert guidance as shown in Algorithm 1. This mechanism aligns the optimization process with expert reasoning, ensuring convergence toward Pareto-optimal solutions that satisfy both system performance metrics and human operational preferences. The integrated process of quantum-enhanced exploration, repository-based exploitation, and expert-guided adaptability is summarized in Algorithm 2.

**Algorithm 1 Pseudo Code for DMiL Update Logic**

    **if** (iteration mod τ == 0):
        **collect expert feedback** (safety, comfort thresholds)
        **update weights:** $w \leftarrow (1 - \gamma) w + \gamma \cdot w\_expert$
        **re-rank Pareto repository** using updated weights

**Algorithm 2 Pseudo Code for QiHSI**

    **Initialize population X**; evaluate objectives; build repository R.
    **for each** iteration t = 1...T do
      **Update control** parameter $c_1 = 2 \exp(-(4t/T)^2)$
      **Select food** source F from R using crowding-based roulette
      **for each salp** i in population:
        **if** i == leader:
          apply quantum superposition and entanglement (Eqs. 1–3)
        **else:**
          update follower chain (Eq. 4)
      **end for**
      **Update repository R;** prune by crowding if |R| > archive_size
      **if** (t mod τ == 0):
        apply DMiL update (Eq. 5)
    **end for**
    **Return Pareto repository R**

Through the integration of quantum superposition, entanglement, and decision-maker feedback, the QiHSI framework enhances both exploration and adaptability without incurring significant computational cost. This results in faster convergence, improved diversity preservation, and superior alignment with real-world calibration objectives critical features for next-generation ADAS systems operating under dynamic and uncertain conditions.

## IV. NUMERICAL EVALUATION AND ANALYSIS

To evaluate the performance and robustness of the proposed Quantum-Inspired Hybrid Swarm Intelligence (QiHSI) framework, comprehensive experiments were conducted on benchmark and real-world multi-objective optimization problems. All algorithms were tested under identical conditions (population size, iteration count, and computational settings) for a fair comparison. Competing methods include MSSA [6], MOPSO [15], SPEA2 [17], MOEA/D [18], NSGA-III [39], and RVEA [40].

### A. Benchmark Evaluation

QiHSI was first validated on standard multi-objective test suites, ZDT and UF which collectively represent convex, discontinuous, and multimodal Pareto front topologies [13], [14]. Each benchmark was executed for 30 independent runs, and results were averaged to ensure statistical reliability. Three well-established performance metrics were employed:

- *Inverted Generational Distance (IGD):* measures proximity of the obtained Pareto front to the true Pareto set (lower is better).
- *Pareto Spread (PSP):* evaluates the uniformity and diversity of solutions across the front (higher is better).
- *Hypervolume (HV):* quantifies the volume of objective space dominated by the Pareto front (higher is better).

Quantitative results as shown in Table 2, demonstrate that QiHSI consistently outperforms all comparison algorithms. It achieves the lowest IGD values (e.g., ZDT1: $7.46 \times 10^{-3}$ vs. MSSA's $1.83 \times 10^{-2}$) and the highest PSP values (e.g., UF4: $2.82 \times 10^{1}$ vs. MSSA's $2.22 \times 10^{1}$), confirming improved convergence and broader front coverage. The HV results further reinforce QiHSI's ability to maintain global exploration while avoiding premature convergence. Despite the inclusion of quantum operations, runtime remained competitive. For instance, QiHSI required 3.87 s per run on UF1 compared to 16.8 s for MOEA/D, validating its computational efficiency in high-dimensional search spaces. Pareto-front visualizations as illustrated in Fig. 1, QiHSI's qualitative performance. On ZDT1, QiHSI achieved near-perfect alignment with the theoretical front, while on ZDT4 (a rugged, multimodal problem), it maintained diversity and avoided local stagnation. Similar dense and uniformly distributed Pareto fronts were observed for UF1 and UF2, confirming QiHSI's balanced exploration–exploitation capability.

### B. Real-World Multi-Criteria ADAS Calibration

QiHSI was next applied to a three-objective ADAS calibration problem involving optimization of:

1. Safety, minimizing collision probability and lateral deviation.
2. Energy Efficiency, minimizing power consumption of sensors and actuators.
3. Driving Comfort, maximizing ride smoothness and reducing jerk.

Performance was quantified using four system-level metrics and two decision-making indicators [37], [38] as shown in Table 1:

TABLE I.
QUALITY INDICATORS FOR ADAS CALIBRATION PROBLEM

| Metric | Explanation | Evaluation Goal |
| --- | --- | --- |
| **Safety Index (SI)** | Aggregate measure of vehicle safety performance derived from sensor accuracy, braking, and response efficiency. | Minimize |
| **Energy Efficiency (EEM)** | Measures fuel consumption and energy optimization by balancing throttle control and sensor efficiency. | Minimize |
| **Comfort Score (CPS)** | Indicates passenger comfort, factoring suspension damping and vehicle response smoothness. | Minimize |
| **System Responsiveness (SR)** | Time taken for the ADAS to respond accurately to varying driving conditions. | Minimize |
| **Decision Convergence (DC)** | Measures stability of decisions after expert feedback, showing reduced fluctuations in solution choices. | High Stability (higher better) |
| **Expert Feedback Alignment (EFA)** | Extent to which obtained solutions match expert preferences after incorporating DMiL feedback. | High Alignment (higher better) |

TABLE III.
QUALITY INDICATORS OF QIHSI COMPARED TO STATE-OF-THE ART ALGORITHMS

| Benchmark Function | Quality Indicator | QiHSI | MSSA | MOPSO | MOEA/D | SPEA2 | NSGA-III | RVEA |
|---|---|---|---|---|---|---|---|---|
| ZDT1 | IGD ↓ | **7.46E-03** | 1.83E-02 | 3.03E+00 | 2.13E-02 | 1.82E-01 | 8.50E-03 | 9.21E-03 |
|  | PSP ↑ | **5.89E+00** | 9.64E+01 | 3.09E-01 | 4.60E+01 | 4.29E+00 | 5.10E+00 | 5.32E+00 |
|  | HV ↑ | **8.70E-01** | 8.20E-01 | 3.09E-01 | 8.50E-01 | 8.23E-01 | 8.61E-01 | 8.58E-01 |
|  | Execution Time ↓ | **5.89E+00** | 6.56E+00 | 1.47E+01 | 1.66E+01 | 1.38E+01 | 6.02E+00 | 6.45E+00 |
| ZDT2 | IGD ↓ | **2.38E-01** | 2.57E-01 | 4.54E+00 | 2.57E-01 | 3.03E-01 | 2.42E-01 | 2.45E-01 |
|  | PSP ↑ | **2.08E+00** | 1.79E+00 | 5.07E-03 | 1.92E+00 | 9.87E-01 | 1.96E+00 | 2.00E+00 |
|  | HV ↑ | **2.15E-01** | 1.68E-01 | 2.16E-01 | 1.92E-01 | 2.03E-01 | 2.11E-01 | 2.09E-01 |
|  | Execution Time ↓ | **6.10E+00** | 6.64E+00 | 1.23E+01 | 1.61E+01 | 1.21E+01 | 6.33E+00 | 6.59E+00 |
| ZDT3 | IGD ↓ | **1.71E-02** | 5.82E-02 | 3.20E+00 | 1.45E-01 | 2.62E-02 | 1.90E-02 | 2.02E-02 |
|  | PSP ↑ | **8.84E+01** | 5.82E-02 | 1.33E-01 | 3.66E+00 | 1.00E+02 | 8.10E+01 | 7.94E+01 |
|  | HV ↑ | **1.49E+00** | 1.21E+00 | 1.32E+00 | 1.06E+00 | 1.31E+00 | 1.42E+00 | 1.39E+00 |
|  | Execution Time ↓ | **3.97E+00** | 4.62E+00 | 1.27E+01 | 1.67E+01 | 1.05E+01 | 4.21E+00 | 4.35E+00 |
| ZDT4 | IGD ↓ | **9.83E-03** | 6.16E-02 | 1.76E+01 | 5.27E-01 | 3.02E-02 | 1.06E-02 | 1.12E-02 |
|  | PSP ↑ | **1.52E+02** | 8.35E+01 | 2.29E-01 | 1.19E+00 | 1.38E+02 | 1.47E+02 | 1.41E+02 |
|  | HV ↑ | **8.75E-01** | 6.83E-01 | 8.64E-01 | 4.08E-02 | 8.71E-01 | 8.74E-01 | 8.72E-01 |
|  | Execution Time ↓ | **5.77E+00** | 6.24E+00 | 1.28E+01 | 1.63E+01 | 1.48E+01 | 6.02E+00 | 6.35E+00 |
| UF1 | IGD ↓ | **2.79E-02** | 4.27E-02 | 1.88E+00 | 1.00E-01 | 8.71E-01 | 3.12E-02 | 3.25E-02 |
|  | PSP ↑ | **4.11E+01** | 4.27E-02 | 5.42E-01 | 8.24E+00 | 1.03E+00 | 3.88E+01 | 3.95E+01 |
|  | HV ↑ | **8.66E-01** | 7.82E-01 | 6.43E-01 | 6.43E-01 | 5.35E-02 | 8.60E-01 | 8.54E-01 |
|  | Execution Time ↓ | **3.87E+00** | 5.90E+00 | 1.31E+01 | 1.68E+01 | 1.51E+01 | 4.10E+00 | 4.32E+00 |
| UF2 | IGD ↓ | **3.87E+00** | 3.54E-02 | 9.91E-01 | 5.03E-02 | 1.65E-02 | 3.64E-02 | 3.72E-02 |
|  | PSP ↑ | **3.65E+01** | 3.32E+01 | 1.00E+00 | 1.61E+01 | 6.97E+01 | 3.41E+01 | 3.50E+01 |
|  | HV ↑ | **8.73E-01** | 8.07E-01 | 8.10E-01 | 8.10E-01 | 8.15E-01 | 8.68E-01 | 8.64E-01 |
|  | Execution Time ↓ | **4.38E+00** | 7.55E+00 | 1.35E+01 | 1.72E+01 | 1.14E+01 | 4.52E+00 | 4.63E+00 |
| UF3 | IGD ↓ | **3.88E-01** | 1.63E-01 | 7.81E+00 | 3.31E-01 | 8.28E-01 | 1.50E-01 | 1.57E-01 |
|  | PSP ↑ | **2.49E+00** | 9.28E+00 | 1.54E-02 | 1.11E+00 | 4.19E-02 | 2.20E+00 | 2.31E+00 |
|  | HV ↑ | **6.78E-01** | 7.03E-01 | 4.66E-01 | 4.15E-01 | 1.60E-01 | 6.65E-01 | 6.70E-01 |
|  | Execution Time ↓ | **3.50E+00** | 5.19E+00 | 8.49E+00 | 1.75E+01 | 1.67E+01 | 3.72E+00 | 3.84E+00 |
| UF4 | IGD ↓ | **3.77E-02** | 4.57E-02 | 2.22E-01 | 7.05E-02 | 2.14E-01 | 3.90E-02 | 3.94E-02 |
|  | PSP ↑ | **2.82E+01** | 2.22E+01 | 2.22E-01 | 1.36E+01 | 6.57E+00 | 2.65E+01 | 2.68E+01 |
|  | HV ↑ | **5.43E-01** | 4.34E-01 | 4.56E-01 | 4.56E-01 | 2.32E-01 | 5.36E-01 | 5.38E-01 |
|  | Execution Time ↓ | **4.80E+00** | 7.05E+00 | 1.48E+01 | 1.72E+01 | 1.30E+01 | 5.02E+00 | 5.18E+00 |

TABLE II.
COMPARATIVE PERFORMANCE OF QIHSI AND OTHER STATE-OF-THE-ART ALGORITHMS ON MULTI-CRITERIA ADAS CALIBRATION

| Algorithm | SI ↓ | EEM ↓ | CPS ↓ | SR ↓ | DC (%) ↑ | EFA (%) ↑ |
|---|---|---|---|---|---|---|
| **QiHSI (+DMiL)** | **0.081** | **0.224** | **0.129** | **0.21** | **94** | **92** |
| MSSA | 0.106 | 0.282 | 0.164 | 0.271 | 74 | 68 |
| MOPSO | 0.118 | 0.293 | 0.175 | 0.28 | 71 | 65 |
| MOEA/D | 0.125 | 0.299 | 0.182 | 0.305 | 68 | 62 |
| SPEA2 | 0.14 | 0.318 | 0.198 | 0.328 | 64 | 59 |
| NSGA-III | 0.095 | 0.25 | 0.142 | 0.232 | 82 | 78 |
| RVEA | 0.099 | 0.263 | 0.148 | 0.241 | 80 | 76 |

- *Safety Index (SI):* aggregated collision risk under varying conditions.
- *Energy Efficiency Metric (EEM):* normalized power utilization score.
- *Comfort Performance Score (CPS):* inverse of acceleration variance.
- *System Responsiveness (SR):* normalized control latency.
- *Decision Convergence (DC):* degree of agreement between algorithmic and expert-guided decisions.
- *Expert Feedback Alignment (EFA):* correlation of optimization outcomes with human expert inputs.

During optimization, domain experts (or a simulated expert model) provided periodic feedback via the DMiL mechanism, enabling real-time adjustment of objective weights based on scenario-specific trade-offs. As summarized in Table 3, QiHSI achieved the best overall calibration quality across all metrics. Specifically, it obtained the lowest SI = 0.081 and EEM = 0.224, indicating superior safety and energy efficiency performance, while maintaining high DC = 94 % and EFA = 92 %, validating DMiL's ability to reflect expert-driven calibration preferences. Competing algorithms exhibited higher safety risk or energy overhead, confirming QiHSI's advantage in managing conflicting objectives through quantum-enhanced adaptability.

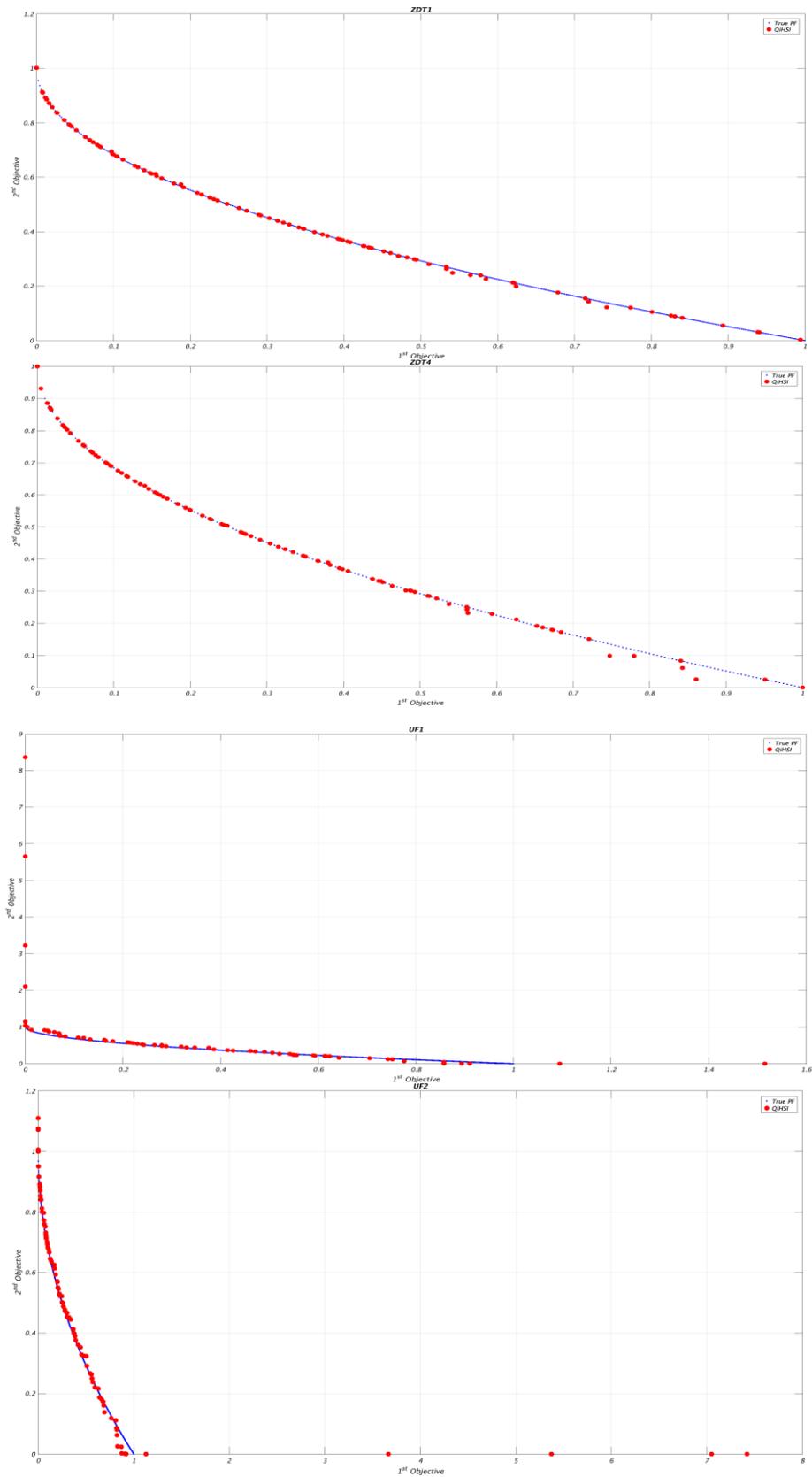

Fig. 1. Qualitative metrics of QiHSI on ZDT1, ZDT4, UF1 & UF2

## C. Complexity and Convergence Analysis

The computational complexity of QiHSI is $O(N \times D \times M)$, where $N$ denotes swarm size, $D$ the problem dimensionality, and $M$ the number of objectives. Empirical runtime analysis over 250 iterations on an Intel i7 (32 GB RAM) system yielded an average execution time of 4.5 s per run, showing near-linear scalability with respect to population size. Convergence plots for IGD and PSP revealed that QiHSI reached stable Pareto fronts faster than MSSA, MOPSO, and MOEA/D, with minimal oscillation across iterations. This confirms that the hybrid quantum-swarm dynamics effectively balance exploration and exploitation, ensuring smooth convergence even under interdependent constraints.

## D. Result Analysis and Discussion

Across all benchmarks and real-world evaluations, QiHSI demonstrated statistically significant improvements (Wilcoxon test, $p < 0.05$) in convergence accuracy and diversity maintenance compared to MSSA, MOPSO, MOEA/D, SPEA2, NSGA-III, and RVEA. The quantum-inspired operators accelerated convergence and preserved diversity, while the DMiL mechanism improved alignment with expert reasoning. Together, these features enable robust, adaptive, and computationally efficient optimization, confirming QiHSI's potential as a practical tool for intelligent ADAS calibration and broader multi-criteria engineering applications.

## V. CONCLUSION

This study presented a Quantum-Inspired Hybrid Swarm Intelligence (QiHSI) framework for multi-criteria ADAS calibration, combining quantum-inspired search mechanisms with swarm intelligence dynamics to enhance exploration, convergence efficiency, and robustness. Through comprehensive benchmarking on standard multi-objective test functions and a real-world ADAS calibration problem, QiHSI consistently outperformed state-of-the-art methods such as MSSA, MOPSO, MOEA/D, and SPEA2 in terms of solution diversity, Pareto front accuracy, and computational efficiency. The integration of the Decision-Maker-in-the-Loop (DMiL) component further enabled adaptive real-time feedback, ensuring the optimization process remains aligned with expert-driven constraints and safety-critical priorities in automotive systems.

Future research will aim to extend QiHSI toward real-time automotive optimization tasks, including dynamic control parameter tuning, adaptive vehicle decision-making, and multi-vehicle coordination. Additionally, the framework will be validated in hardware-in-the-loop (HIL) and vehicle-in-the-loop (VIL) environments to assess its operational reliability under realistic sensing, latency, and environmental constraints, ultimately advancing intelligent and self-adaptive optimization in autonomous transportation systems.

## DECLARATION ON GENERATIVE AI

During the preparation of this article, the author(s) used ChatGPT in order to check grammar, spelling and improving clarity of expression. The author(s) take(s) full responsibility for the article's content.